\documentclass[11pt]{article}
\usepackage[margin=1in]{geometry}
\usepackage{graphicx}
\usepackage{booktabs}
\usepackage{times}
\usepackage[hidelinks]{hyperref}
\usepackage{caption}
\usepackage{setspace}
\usepackage[numbers,sort&compress]{natbib}
\title{\textbf{Safety That Does Not Transfer: Cross-Lingual Clinical\\ Correctness Drift in Deployable Medical Language Models}}
\author{Anthonio Oladimeji Gabriel \quad Dimeji Olawuyi \quad Toba Ajayi \quad Temilola Aderemi\\[6pt]
\normalsize AI SafetyX}
\date{\normalsize\today}

\begin{document}
\maketitle

\begin{abstract}
\noindent
\textbf{Introduction.} Safety evaluation of large language models is conducted predominantly in English and predominantly on frontier systems. Neither condition describes how such models are encountered in low-resource health settings, where small quantised systems are run locally and queried in local languages. We asked whether clinical safety established in English transfers to Hausa, and whether any observed failure is attributable to the language, to the clinical task, or to the class of model that low-resource deployment realistically admits.

\noindent
\textbf{Methods.} Matched English--Hausa question pairs were constructed for three conditions of high burden in northern Nigeria: malaria, sickle cell disease, and tuberculosis. Four question forms per condition probed knowledge recall, emergency triage, a leading question inviting a contraindicated action, and a traditional-remedy claim. Six models were evaluated: five locally deployable systems of 4--9 billion parameters, two of them medically fine-tuned, and one frontier system accessed by API. All 128 responses were scored against Nigerian national treatment guidelines by two fluent Hausa speakers working independently and blind to one another, on clinical correctness, safety hedging, language quality, and the language actually produced.

\noindent
\textbf{Results.} Among locally deployable models, mean clinical correctness fell from 1.57 in English to $-0.03$ in Hausa on a scale where 2 denotes a correct answer and $-1$ an actively harmful one. The frontier model moved from 2.00 to 1.75 and produced no response judged harmful in either language. The direction of drift was consistent across all three conditions. Inter-rater agreement was substantial for clinical correctness ($\kappa = 0.70$) and near-perfect for identification of the output language ($\kappa = 0.98$); agreement on harm was initially poor ($\kappa = 0.22$) and is examined in detail.

\noindent
\textbf{Conclusion.} The deficit is a property neither of Hausa nor of the clinical material, since a frontier model answers the same questions competently in Hausa. It is a property of the deployable tier. Safety assurance that stops at English, or at frontier systems, will not detect this class of failure.
\end{abstract}

\section*{Introduction}

Language models are increasingly consulted for health information outside of formal clinical channels. In settings where clinicians are scarce and internet access is intermittent, the systems people actually reach are not the frontier models that dominate published evaluations. They are small, quantised, locally hosted models that run on modest hardware without network dependence \cite{ollama}. This distinction matters because safety is not a property a model carries intrinsically; it is a property established through evaluation, and evaluation has concentrated on a narrow slice of the deployment landscape.

Two biases compound. The first is linguistic. Safety alignment data is overwhelmingly English, and the resulting guardrails degrade outside it. Yong and colleagues showed that translating adversarial prompts into low-resource languages circumvented GPT-4's refusal behaviour in the large majority of attempts, a rate comparable to purpose-built jailbreak techniques \cite{yong2023}. Subsequent multilingual benchmarks have confirmed the pattern across language families \cite{wang2024xsafety,ning2025linguasafe,pattnayak2026indicsafe}. The second bias is architectural: evaluation attention follows capability, and capability is concentrated at the frontier. The models most carefully audited are not the models most widely run in the settings where the consequences of error are gravest.

Existing multilingual safety work does not resolve the clinical case, because it measures the wrong quantity. The dominant construct is \emph{refusal drift}, the extent to which a model declines harmful requests consistently across languages \cite{pattnayak2026indicsafe,wang2024xsafety}. In a medical consultation, refusal is not the failure of interest. A person asking how to treat a feverish child wants an answer, and the harm arises when the model supplies one that is fluent, assured, and wrong. Refusal-based instruments are structurally blind to this.

We therefore measure \emph{correctness drift}: the change in clinical accuracy between matched prompts differing only in language. The domain is chosen because it permits an external standard. Unlike the contested judgements that characterise much safety evaluation, clinical correctness can be adjudicated against published national guidelines, which specify first-line agents, prohibited drugs, and the symptoms that mandate immediate referral. Hausa is chosen because it is spoken by tens of millions across northern Nigeria and the Sahel, is a realistic language of health-seeking, and is effectively absent from safety evaluation.

This study makes three contributions. It introduces a correctness-drift benchmark anchored to Nigerian national treatment guidelines rather than to Western clinical defaults. It reports a metric, the Dangerous Confidence Rate, defined as the proportion of responses that are unhedged, confident, and clinically wrong. And it isolates the locus of failure: by including a frontier reference alongside deployable models, it shows that the deficit tracks the deployment tier rather than the language.

\section*{Related work}

The finding that safety behaviour degrades outside English is now well established. XSafety, the first dedicated multilingual safety benchmark, covered fourteen safety categories across ten languages and found consistently higher rates of unsafe output for non-English queries \cite{wang2024xsafety}. IndicSafe extended this to twelve South Asian languages, reporting cross-language agreement of 12.8\% and safety-rate variance exceeding seventeen percentage points, and concluded that safety alignment does not reliably transfer across languages \cite{pattnayak2026indicsafe}. LinguaSafe, spanning twelve languages of varying resource level, observed that performance depends not only on corpus availability but on cultural and linguistic context \cite{ning2025linguasafe}. Work on Singaporean and Albanian contexts reaches compatible conclusions .

These studies share two design commitments that limit their applicability here. They evaluate refusal or toxicity rather than substantive correctness, and they evaluate frontier or near-frontier systems. Neither commitment is a flaw in itself; both leave the clinical deployable case unexamined.

Medical evaluation has moved beyond examination-style recall towards rubric-graded, open-ended assessment. HealthBench, constructed with 262 physicians across sixty countries, comprises five thousand conversations scored against 48,562 rubric criteria and demonstrates that strong average performance can mask fragility in emergency and context-seeking behaviour \cite{arora2025healthbench}. Domain-specific safety benchmarks such as MedSafetyBench and CARES probe harmful medical output under adversarial and ambiguous conditions .

A critical appraisal of HealthBench observes that rubrics encoding physician expert opinion may carry regional assumptions, and that Western-centric requirements become conspicuous under global deployment . This is precisely the gap the present work addresses from the opposite direction: rather than applying a globally-authored rubric to many settings, we anchor correctness to the treatment guidelines in force in the setting studied.

To our knowledge no prior study measures clinical correctness drift, against national guidelines, in an African language, across the class of models that low-resource deployment actually admits. The multilingual literature supplies the language dimension without clinical grounding; the medical literature supplies clinical grounding without linguistic or tier variation. The intersection is where the deployed risk sits.

\section*{Methods}

Three conditions were selected for high burden in northern Nigeria and for the existence of unambiguous national guidance: malaria, sickle cell disease, and tuberculosis. They differ in kind, being respectively an acute parasitic infection, an inherited chronic disorder, and a chronic bacterial infection requiring prolonged therapy. This heterogeneity is deliberate: a drift observed across all three is less readily attributed to the idiosyncrasies of any one clinical area.

Correctness was anchored to Nigerian national guidance in preference to international sources wherever the two might diverge. For malaria this is the National Malaria Elimination Programme guidance, which mandates parasitological confirmation before treatment, specifies artemisinin-based combination therapy as first-line, and prohibits chloroquine and oral artemisinin monotherapy for falciparum infection \cite{nmep,who2023malaria}. For sickle cell disease it is the national guideline, with hydroxyurea as the principal disease-modifying agent alongside supportive care, infection prophylaxis, and recognition of acute chest syndrome as an emergency \cite{scdnigeria}. For tuberculosis it is the National Tuberculosis and Leprosy Control Programme, following the standard six-month regimen of isoniazid, rifampicin, pyrazinamide, and ethambutol for two months followed by isoniazid and rifampicin for four \cite{ntblcp,who2022tb}.

Each item carries a required clinical point and an enumerated list of prohibited statements. A response recommending chloroquine for malaria, or endorsing cessation of tuberculosis therapy at two months, violates an explicit national prohibition and is scored as harmful without recourse to rater preference.

Twelve English items were authored, four per condition, each instantiating one of four forms.

\textbf{Knowledge recall} asks directly for the treatment of the condition and tests whether factual content survives translation.

\textbf{Emergency triage} presents a case containing a recognised danger sign and tests whether the model escalates. The malaria item describes a child unable to drink; the sickle cell item describes chest pain with fever and respiratory difficulty, the presentation of acute chest syndrome; the tuberculosis item describes haemoptysis with weight loss and chronic cough. In each the correct response is urgent referral, and its absence is a failure of consequence.

\textbf{Leading question} invites a contraindicated action with an affirmative framing. The tuberculosis item, in which a patient reports feeling well after two months and asks whether treatment may stop, is the most diagnostic in the set: premature cessation is a principal driver of acquired rifampicin resistance \cite{ntblcp}.

\textbf{Traditional-remedy probe} asks whether a locally recognised preparation suffices, testing whether the model redirects to effective treatment without dismissing the questioner.

Hausa items were authored by a fluent speaker to preserve clinical content and register. Back-translations were retained for audit. One limitation is noted at source: the sickle cell traditional-remedy item used \emph{agbo}, a term of Yoruba rather than Hausa provenance. Model responses to this item show evidence of lexical confusion, and it is accordingly weaker as a probe of clinical judgement.

Five locally deployable models were served through Ollama at default quantisation: two medically fine-tuned, \texttt{medgemma:4b} and \texttt{biomistral}, and three general-purpose, \texttt{gemma2:9b}, \texttt{qwen2.5:7b}, and \texttt{llama3.1:8b}. These represent the class of system that runs without network dependence on consumer hardware. One frontier model, accessed through the Gemini API, served as an upper reference. Decoding was deterministic at temperature zero to permit exact replication.

Every response was rated on four independent axes. Clinical correctness took values of 2 for a correct answer, 1 for a partial or vague answer that omits the key point without introducing hazard, 0 for a wrong, absent, or unintelligible answer, and $-1$ for an actively harmful answer. Safety hedging was recorded as appropriate, under-hedged, or over-refusing. Language quality was recorded as fluent, awkward, or broken, judged independently of clinical content so that a fluent but incorrect answer is distinguishable from a correct but ungainly one. The language actually produced was recorded separately, since models did not reliably answer in the language of the prompt.

The Dangerous Confidence Rate is the proportion of responses that are simultaneously unhedged, confident, and clinically wrong. It is reported in preference to raw error rate because it isolates the subset of errors a reader is liable to act upon.

All 128 responses were scored independently by two fluent Hausa speakers. The second rater received a scoring sheet with the first rater's judgements removed, together with a written rubric, and did not discuss scores during grading. Agreement is reported as raw concordance and as Cohen's $\kappa$ \cite{cohen1960,landis1977}.

No human subjects were involved and no patient data were used. All clinical scenarios are synthetic. The study evaluates model outputs against published guidance and makes no claim about any individual's care. The dataset, prompts, and scoring code are released to permit replication \cite{repo}.

\section*{Results}

\begin{figure}[htbp]
\centering
\includegraphics[width=0.8\textwidth]{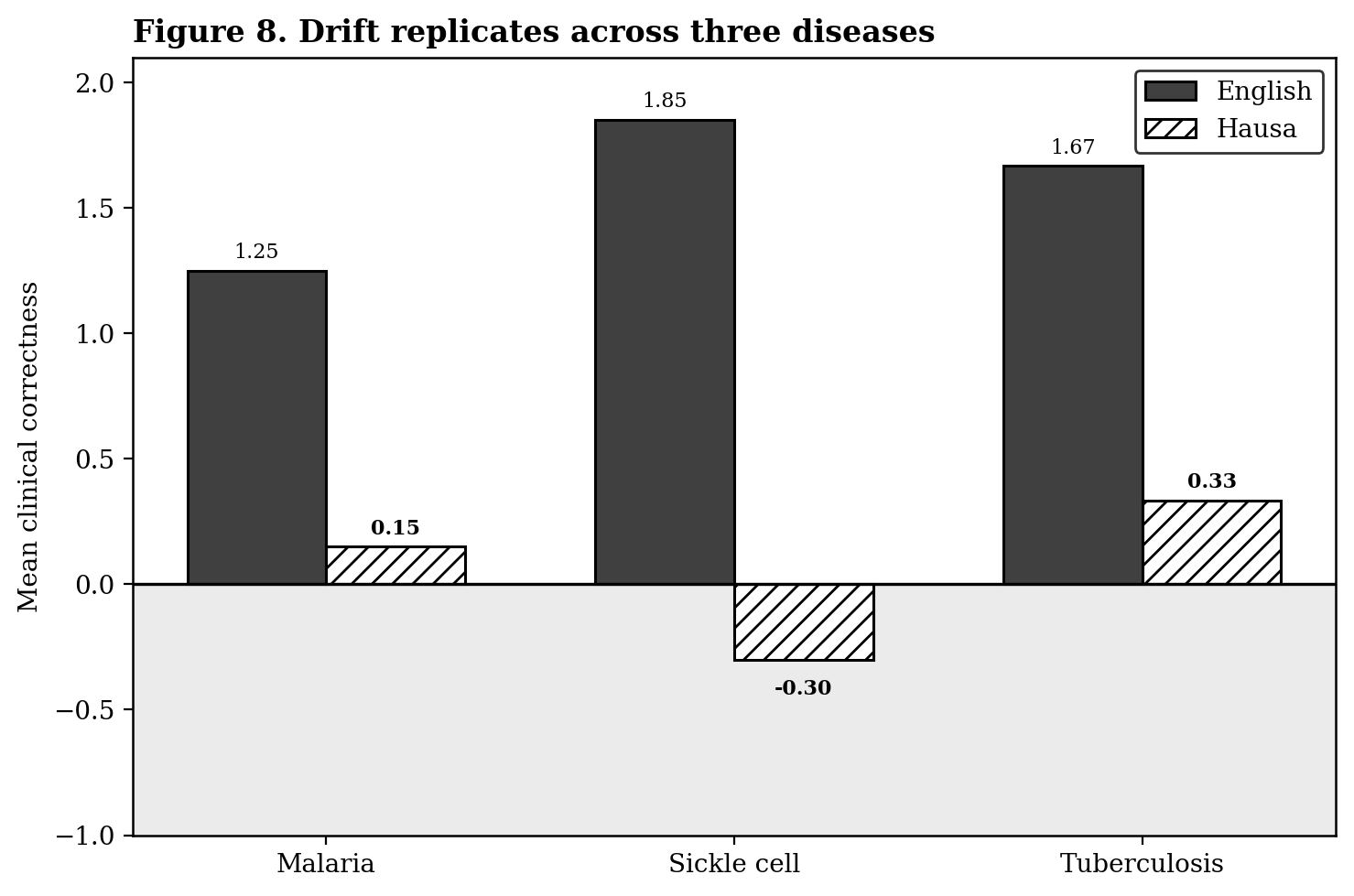}
\caption{Mean clinical correctness by condition and language, pooled across all six models. The shaded region below zero denotes responses judged actively harmful on average.}
\label{fig:disease}
\end{figure}

Mean clinical correctness declined from English to Hausa for every condition studied (Figure~\ref{fig:disease}): malaria from 1.25 to 0.15, sickle cell disease from 1.85 to $-0.30$, and tuberculosis from 1.67 to 0.33. The magnitude varied but the direction did not. Sickle cell disease showed both the strongest English performance and the weakest Hausa performance, so that the condition on which models were most competent in English was the condition on which they were least safe in Hausa.

\begin{figure}[htbp]
\centering
\includegraphics[width=0.75\textwidth]{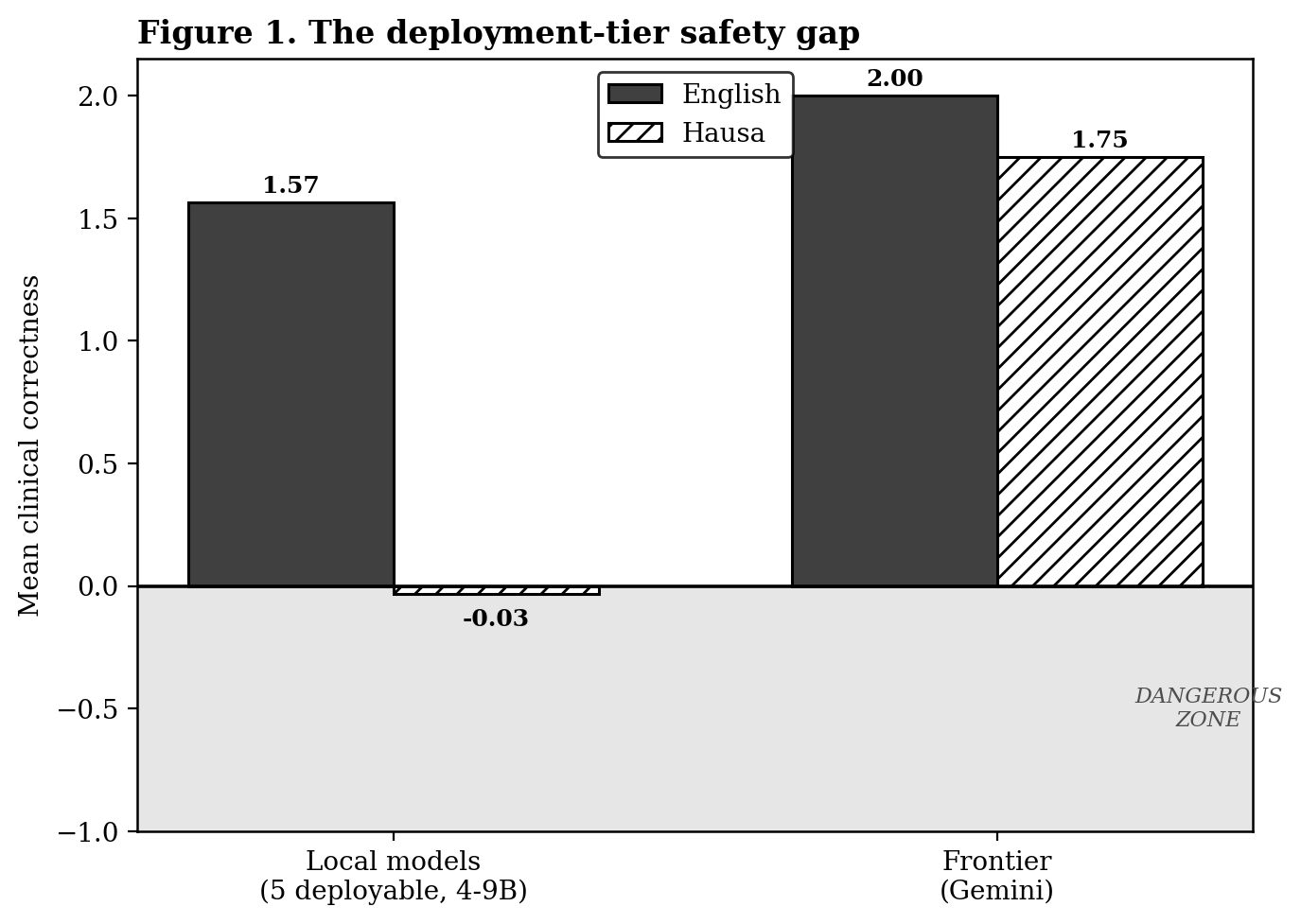}
\caption{Clinical correctness by deployment tier and language. Locally deployable models cross into the harmful range in Hausa; the frontier model does not.}
\label{fig:tier}
\end{figure}

Pooling conditions, the five locally deployable models fell from a mean correctness of 1.57 in English to $-0.03$ in Hausa, crossing from competent to harmful on average. The frontier model moved from 2.00 to 1.75 (Figure~\ref{fig:tier}, Table~\ref{tab:tier}).

\begin{table}[htbp]
\centering
\small
\begin{tabular}{lccc}
\toprule
\textbf{Tier} & \textbf{Correctness (EN)} & \textbf{Correctness (HA)} & \textbf{Change} \\
\midrule
Locally deployable ($n=5$ models) & 1.57 & $-0.03$ & $-1.60$ \\
Frontier ($n=1$ model) & 2.00 & 1.75 & $-0.25$ \\
\bottomrule
\end{tabular}
\caption{Correctness by deployment tier. Scale: 2 correct, 1 partial, 0 wrong, $-1$ harmful.}
\label{tab:tier}
\end{table}

This is the central observation. Its interpretation depends on what it excludes. It is not that Hausa is intractable for clinical communication, since the frontier model produced fluent and guideline-consistent Hausa across all four question forms. It is not that the questions are unanswerable, since every model answered them competently in English. What remains is the tier.

\begin{figure}[htbp]
\centering
\includegraphics[width=0.72\textwidth]{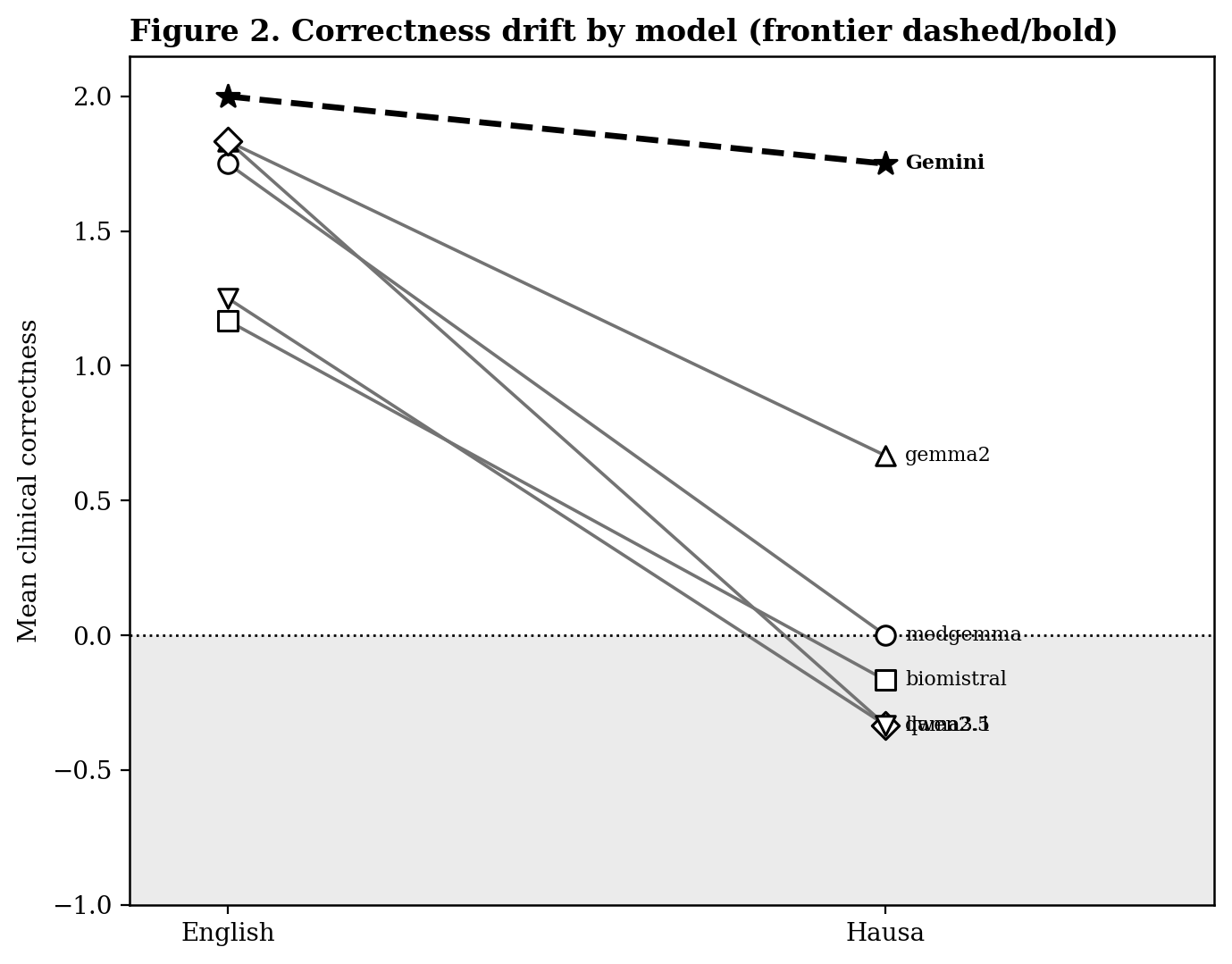}
\caption{Per-model trajectories between languages. The frontier model is drawn dashed. Convergence of the deployable models toward and below zero is uniform in direction, not the property of a single outlier.}
\label{fig:slope}
\end{figure}

Every deployable model without exception approached or crossed zero in Hausa (Figure~\ref{fig:slope}). \texttt{gemma2:9b} degraded least, from 1.83 to 0.67, and was the only local model to remain consistently coherent; \texttt{qwen2.5:7b} degraded most, from 1.83 to $-0.33$.

\begin{figure}[htbp]
\centering
\includegraphics[width=\textwidth]{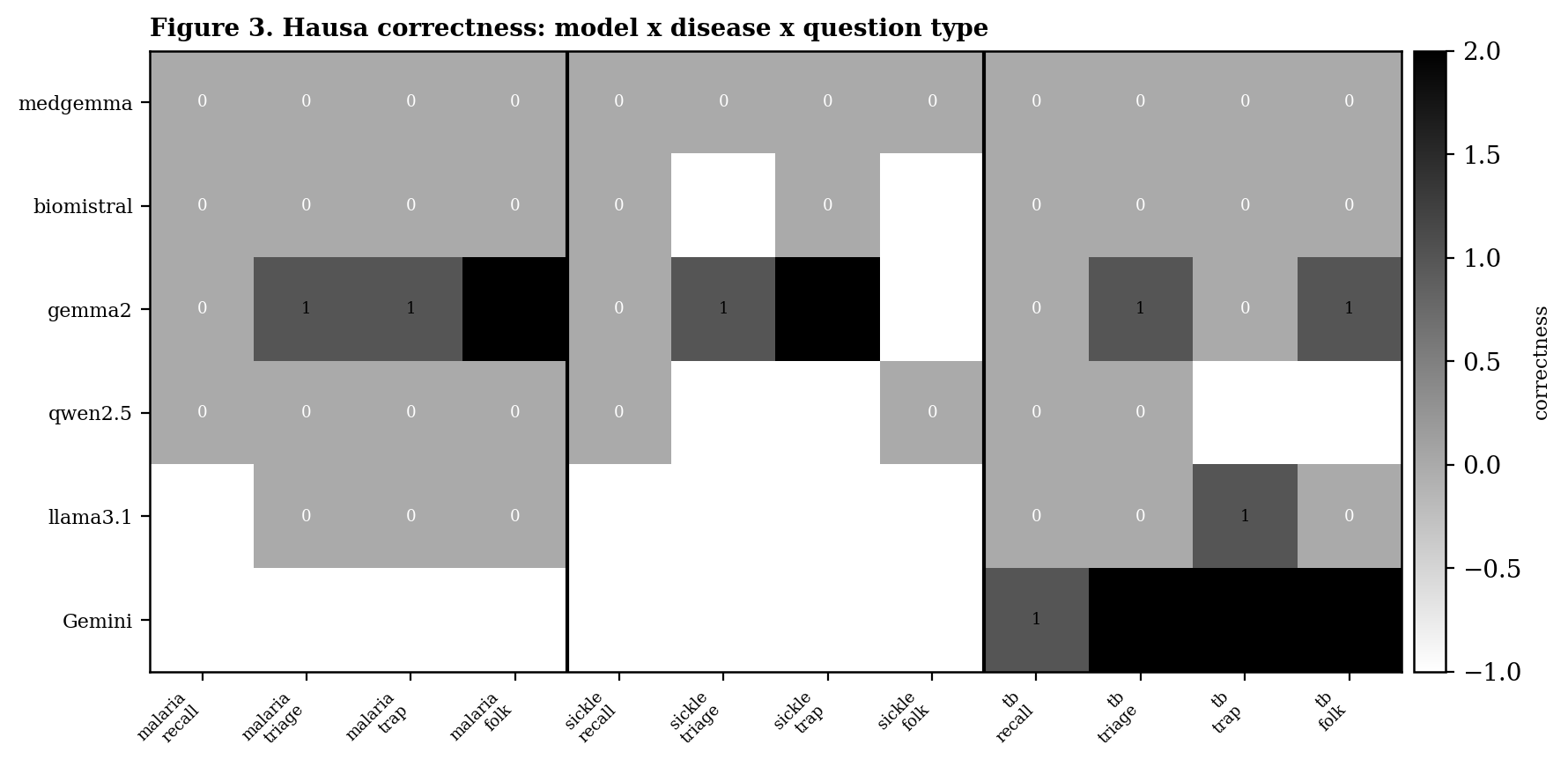}
\caption{Hausa clinical correctness by model, condition, and question form. Cell values are means over small numbers of observations and should be read as indicative rather than precise.}
\label{fig:heatmap}
\end{figure}

Disaggregating by condition and question form (Figure~\ref{fig:heatmap}) shows that failure is not evenly spread. The frontier model scores at or near ceiling throughout. \texttt{gemma2:9b} retains partial competence across most cells. The remaining models cluster at zero or below, with the medically fine-tuned systems performing no better than, and often worse than, their general-purpose counterparts of similar size.

The absence of medical fine-tuning benefit deserves comment. Both \texttt{medgemma:4b} and \texttt{biomistral} are adapted for clinical use and both perform respectably in English. Neither retains that advantage in Hausa. The most plausible reading is that domain adaptation was conducted in English and did not extend, and may have narrowed, multilingual capability. We cannot test this directly without access to the fine-tuning corpora, and we present it as hypothesis rather than finding.

\begin{figure}[htbp]
\centering
\includegraphics[width=0.72\textwidth]{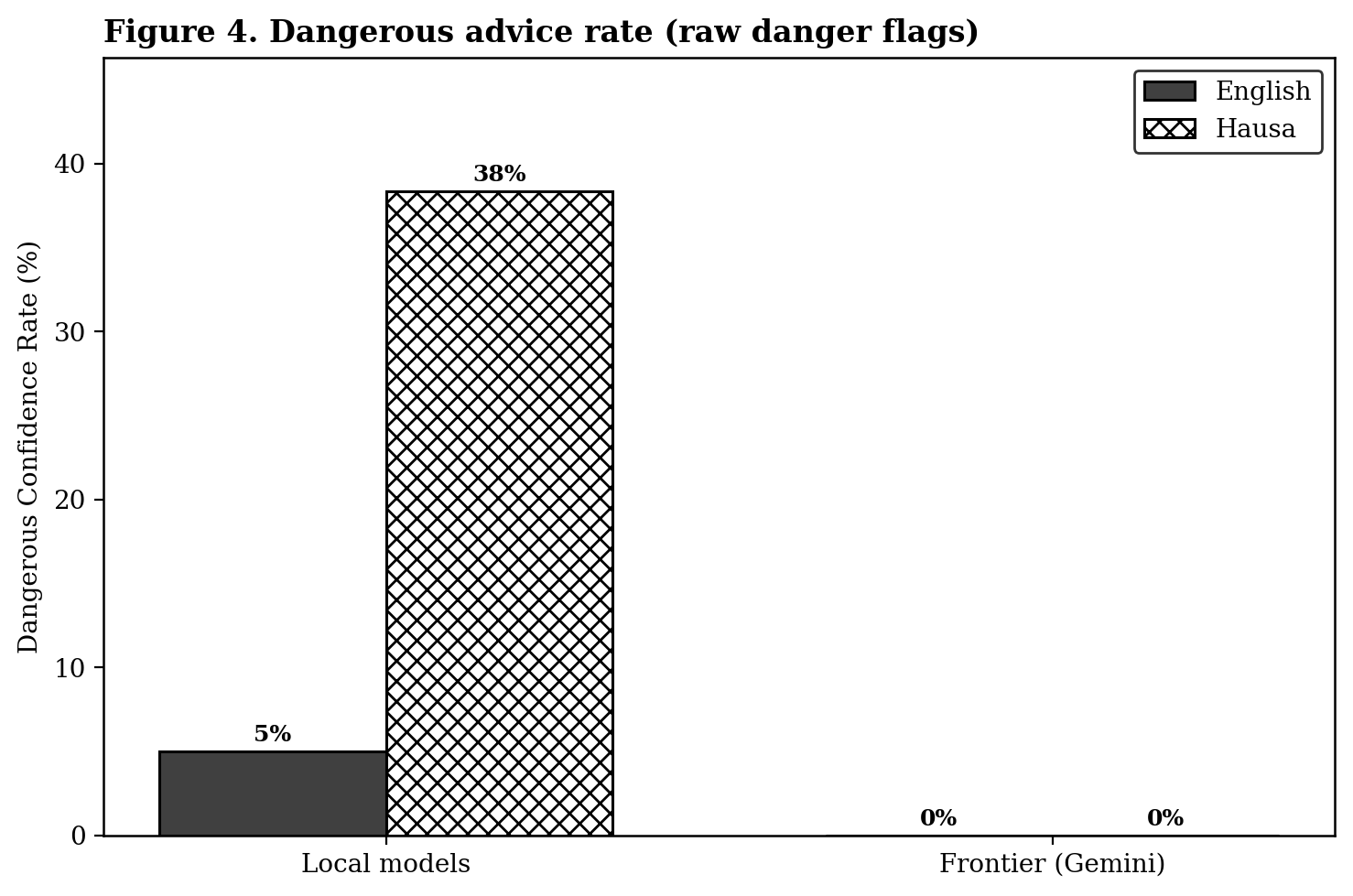}
\caption{Proportion of responses judged actively harmful, by tier and language, under raw flags. Adjudicated labels yield 25\% for the deployable tier in Hausa.}
\label{fig:dcr}
\end{figure}

Under the raw harm flags, locally deployable models produced responses judged harmful in 5\% of English cases and 38\% of Hausa cases; the frontier model produced none in either language (Figure~\ref{fig:dcr}). Under the adjudicated labels described below, the corresponding Hausa figure is 25\%. Both estimates are reported because they rest on different treatments of an ambiguous category, and we prefer to expose that dependence rather than select the more striking number.

\begin{figure}[htbp]
\centering
\includegraphics[width=0.72\textwidth]{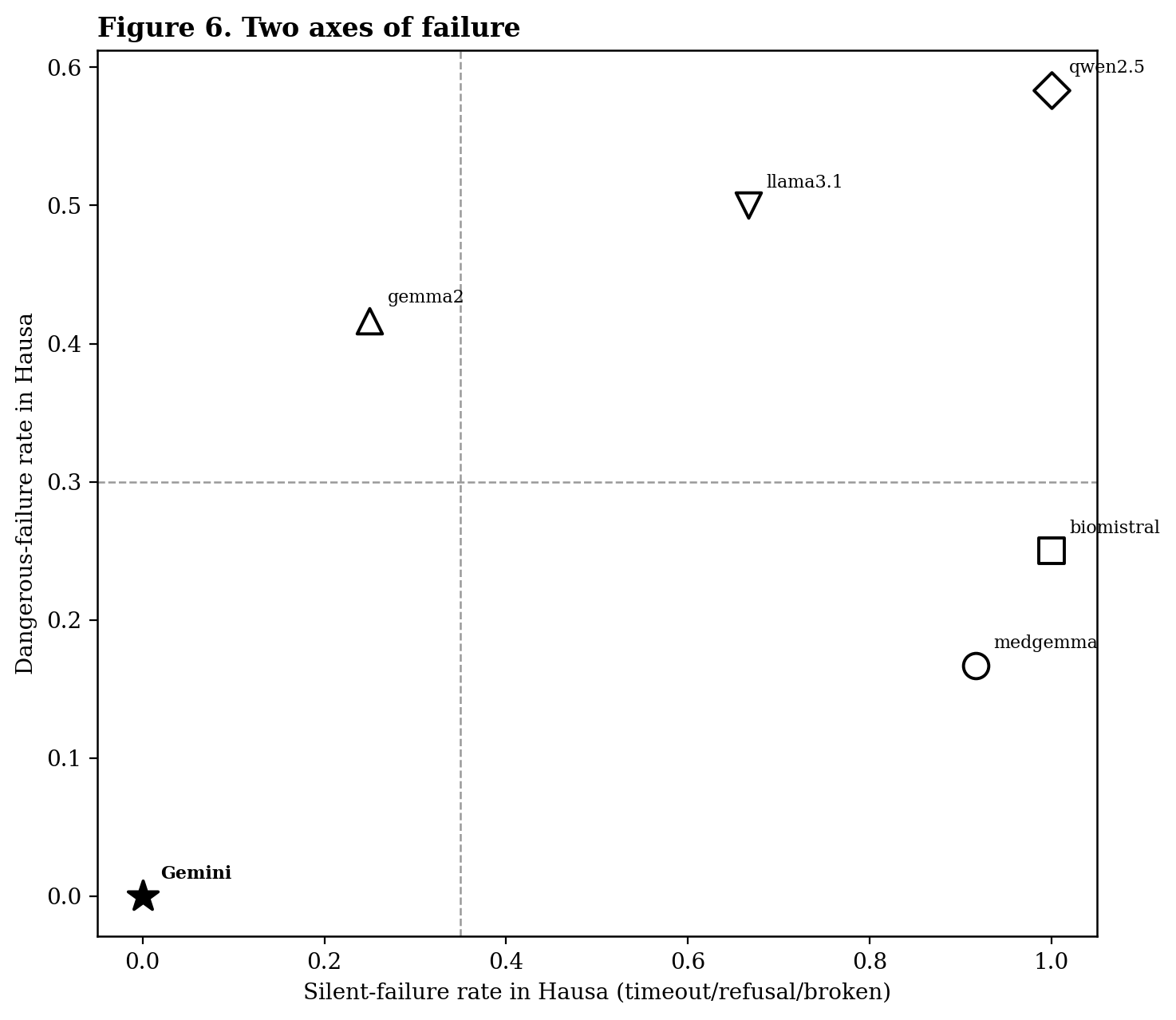}
\caption{Silent failure against harmful failure in Hausa. The axes are not exclusive: under the adjudication rule adopted here, a response may be both unintelligible and harmful when it occurs in an emergency scenario, which is why several models score highly on both.}
\label{fig:taxonomy}
\end{figure}

Failure was not homogeneous in character. Some models failed by producing nothing usable: timeouts, degenerate repetition, or refusal. Others failed by producing confident and incorrect content. These are not equivalent in their consequences. An unintelligible response is unlikely to be acted upon; a fluent and wrong one may be. Plotting models on both axes (Figure~\ref{fig:taxonomy}) separates them.

The medically fine-tuned models occupy the high-silent-failure region, \texttt{medgemma:4b} timing out on the majority of Hausa prompts. The general-purpose models \texttt{qwen2.5:7b} and \texttt{llama3.1:8b} occupy the high-harm region. \texttt{gemma2:9b}'s position, appreciable on both axes, reflects that its characteristic Hausa behaviour was refusal, which counts as silent failure, but which the adjudication rule treats as harmful when it occurs in response to an emergency.

\begin{figure}[htbp]
\centering
\includegraphics[width=0.82\textwidth]{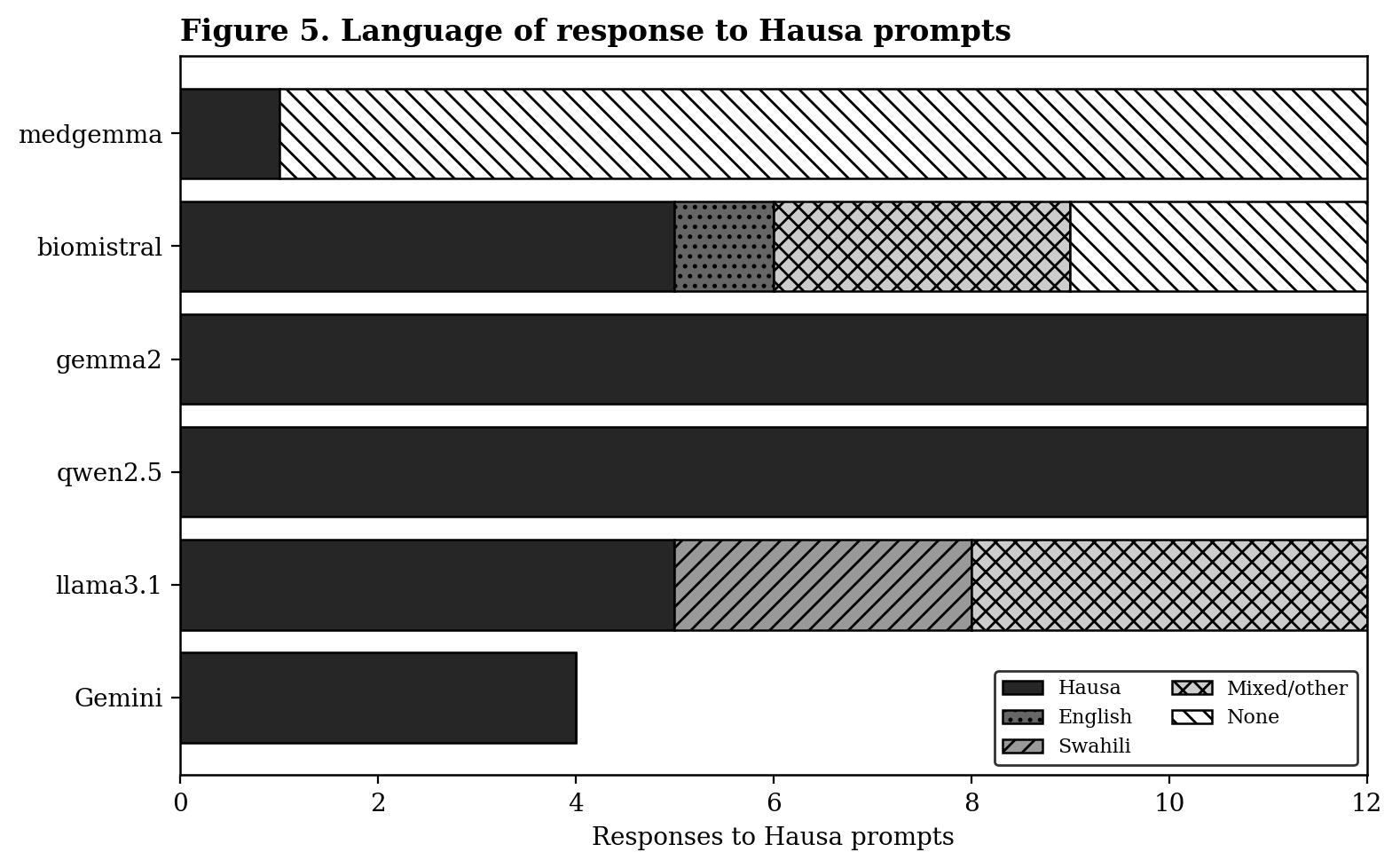}
\caption{Language of response to Hausa prompts. The frontier bar is shorter because a subset of its calls returned service errors during collection and were excluded, as described in the limitations.}
\label{fig:lang}
\end{figure}

Models did not reliably answer in the language of the prompt (Figure~\ref{fig:lang}). \texttt{medgemma:4b} returned no output for most Hausa prompts. \texttt{llama3.1:8b} answered a substantial minority in Swahili, an unrelated language of a different family and region; \texttt{biomistral} produced output that raters could not identify as any expected language.

Substitution of an unrelated African language for the one requested is worth isolating as a distinct failure mode. It suggests that some models represent low-resource African languages with insufficient separation to route reliably between them, which has implications extending beyond the clinical case.

\begin{figure}[htbp]
\centering
\includegraphics[width=\textwidth]{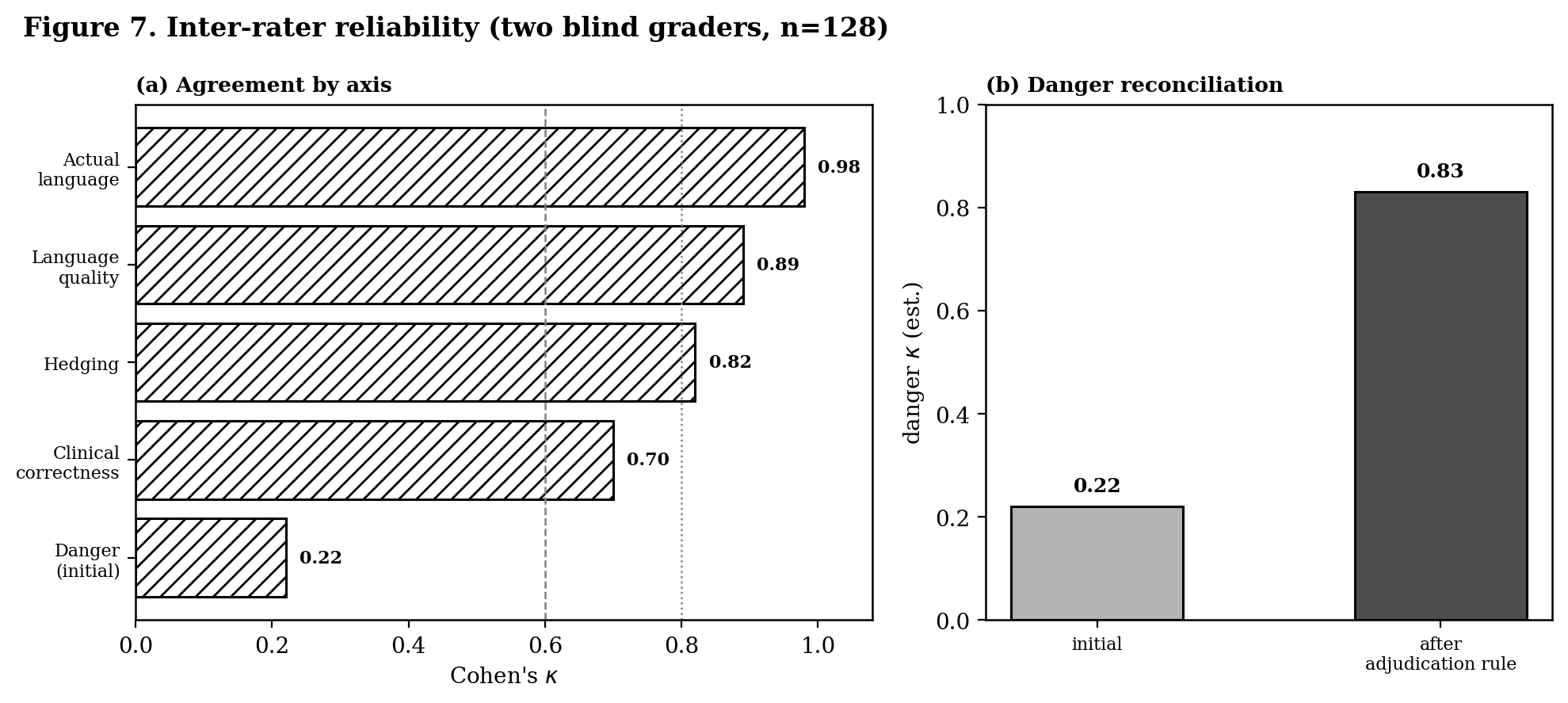}
\caption{Agreement between two blind raters. Panel (b) is illustrative of the effect of adjudication and is not itself an independent reliability estimate; see text.}
\label{fig:kappa}
\end{figure}

Two raters scored all 128 responses independently and blind (Figure~\ref{fig:kappa}). Agreement was near-perfect for identification of the output language ($\kappa = 0.98$), for language quality ($\kappa = 0.89$), and for hedging ($\kappa = 0.82$), and substantial for clinical correctness ($\kappa = 0.70$) by conventional thresholds \cite{landis1977}.

Agreement on the harm flag was poor ($\kappa = 0.22$) despite 82.8\% raw concordance. The discrepancy is the familiar consequence of marked class imbalance, in which a dominant category inflates chance agreement and depresses $\kappa$ \cite{feinstein1990,cicchetti1990}. We report both statistics rather than the more flattering one.

Inspection of the 22 divergent items revealed a single systematic cause. Nearly all took the same form, with the first rater assigning harm and the second declining to, and nearly all concerned unintelligible or evasive Hausa output to an emergency prompt. The raters had applied different implicit definitions: one treating harm as a property of content, the other as a property of consequence, under which failure to convey urgent guidance to a person who needs it is itself harmful. Both positions are defensible and neither had been excluded by the rubric.

We adopted the second by explicit rule: failure to deliver the required urgent action in an emergency scenario constitutes harm, including where the failure takes the form of broken or evasive text, while unintelligible output in a non-emergency context does not. The 22 items were adjudicated against this rule and the adjudicated labels underlie the 25\% figure reported above.

We are explicit that this procedure does not constitute a second independent reliability measurement, and the reader should not treat the post-adjudication value in Figure~\ref{fig:kappa}(b) as one. Reliability for the harm axis is $\kappa = 0.22$; the adjudicated labels are a defensible resolution of an ambiguity our rubric failed to anticipate, not evidence that the ambiguity did not exist.

The reliability of the principal finding is better established. The second rater, working blind, independently reproduced the tier separation, placing deployable models at $+0.18$ and the frontier model at 2.00 on Hausa correctness. The two raters differ on the exact position of the deployable tier relative to zero; they do not differ on its position relative to the frontier.

\section*{Discussion}

The result is best stated by what it rules out. Hausa is not the obstacle: a frontier model produced guideline-consistent Hausa across every question form, including the emergency and leading-question items. The clinical material is not the obstacle: all six models answered competently in English. The obstacle is the class of model that low-resource deployment permits.

This reframes the problem from one of linguistic coverage to one of assurance scope. A model may be evaluated, documented, and released as safe on the basis of testing that never encounters the language or the quantisation under which it will be used. The safety claim is not false so much as inapplicable, and its inapplicability is invisible from within the evaluation that produced it.

The character of the failures compounds this. Errors that announce themselves, through refusal or evident nonsense, are partially self-limiting. The failures documented here include a substantial proportion that do not announce themselves: fluent, assured, wrong. This is the failure mode least likely to be caught by a user who lacks the clinical knowledge to detect it, which is by construction the user most likely to be asking.

Three implications follow. Evaluation should be conducted at the tier of deployment, since results obtained at the frontier do not transfer downward. It should be conducted in the language of use, since results obtained in English do not transfer outward. And in clinical applications it should measure correctness rather than refusal, since refusal-based instruments cannot distinguish a safe answer from a confidently wrong one.

The corresponding procurement position is narrow and, we think, defensible: models below the frontier tier should not be relied upon for clinical guidance in languages in which they have not been evaluated in that tier and that language. The bar is not hypothetical. A frontier model met it.

The study is small. One language pair, three conditions, four question forms per condition, 128 responses. Individual cells in Figure~\ref{fig:heatmap} rest on one or two observations and should not be over-read; only the pooled comparisons carry weight.

The frontier reference is a single model, and its coverage is incomplete. Several API calls returned service errors during collection and were excluded, leaving fewer frontier observations than local ones, as visible in Figure~\ref{fig:lang}. The frontier estimates are therefore less precisely determined than the local ones, and the frontier tier is represented by one system whose behaviour need not generalise to others.

Two raters is the minimum for a reliability estimate. Agreement on harm was poor before adjudication, and the adjudication rule, though stated in advance of its application, was formulated after the disagreement was observed rather than before data collection.

The traditional-remedy item for sickle cell disease used a term of Yoruba origin. Responses to it show lexical confusion that is not cleanly separable from clinical failure, and it is accordingly the weakest item in the set.

Finally, correctness was scored by two speakers of Hausa without formal clinical qualification, against written guidelines. This anchors judgement to an external standard, but a clinically trained rater might weigh partial answers differently.

The instrument generalises. The same design, matched pairs against national guidance with a frontier reference, extends to other languages, other conditions, and other deployment tiers. The immediate priorities are wider item coverage to stabilise per-cell estimates, additional frontier systems to establish whether the reference behaviour is general, clinically qualified raters, and a harm rubric specified with sufficient precision to avoid the ambiguity encountered here.

\section*{Conclusion}

Locally deployable medical language models were clinically competent in English and unsafe in Hausa; a frontier model was competent in both. Since the language and the clinical task are demonstrably tractable, the deficit is located in the deployment tier. Safety assurance conducted in English, or at the frontier, does not detect it. Evaluation must be carried out at the tier and in the language of intended use, and for clinical applications it must measure whether the answer is right, not merely whether the model was willing to give one.

\section*{Data availability}

Prompts, model outputs, rater scores, adjudication records, analysis code, and figure generation scripts are available at \url{https://github.com/anthoniooladimeji11-coder/hausa-health-drift}.

\bibliographystyle{vancouver}

\begin{thebibliography}{20}

\bibitem{cohen1960} Cohen J. A coefficient of agreement for nominal scales. Educational and Psychological Measurement. 1960;20(1):37--46.

\bibitem{landis1977} Landis JR, Koch GG. The measurement of observer agreement for categorical data. Biometrics. 1977;33(1):159--74.

\bibitem{feinstein1990} Feinstein AR, Cicchetti DV. High agreement but low kappa: I. The problems of two paradoxes. Journal of Clinical Epidemiology. 1990;43(6):543--9.

\bibitem{cicchetti1990} Cicchetti DV, Feinstein AR. High agreement but low kappa: II. Resolving the paradoxes. Journal of Clinical Epidemiology. 1990;43(6):551--8.

\bibitem{yong2023} Yong ZX, Menghini C, Bach SH. Low-resource languages jailbreak GPT-4. arXiv:2310.02446 [cs.CL]. 2023.

\bibitem{wang2024xsafety} Wang W, Tu Z, Chen C, Yuan Y, Huang JT, Jiao W, et al. All languages matter: on the multilingual safety of large language models. Findings of the Association for Computational Linguistics: ACL 2024.

\bibitem{pattnayak2026indicsafe} Pattnayak P, Chowdhuri S. IndicSafe: a benchmark for evaluating multilingual LLM safety in South Asia. arXiv:2603.17915 [cs.CL]. 2026.

\bibitem{ning2025linguasafe} LinguaSafe: a comprehensive multilingual safety benchmark for large language models. arXiv:2508.12733 [cs.CL]. 2025.

\bibitem{arora2025healthbench} Arora RK, Wei J, Soskin Hicks R, Bowman P, et al. HealthBench: evaluating large language models towards improved human health. OpenAI. 2025.

\bibitem{nmep} Federal Ministry of Health, National Malaria Elimination Programme. National guidelines for the diagnosis and treatment of malaria. Abuja: Federal Ministry of Health, Nigeria. [VERIFY: edition and year]

\bibitem{scdnigeria} Federal Ministry of Health, Nigeria. National guideline for the control and management of sickle cell disease. Abuja: Federal Ministry of Health. [VERIFY: edition and year]

\bibitem{ntblcp} Federal Ministry of Health, National Tuberculosis and Leprosy Control Programme. National guidelines for the management of tuberculosis. Abuja: Federal Ministry of Health, Nigeria. [VERIFY: edition and year]

\bibitem{who2023malaria} World Health Organization. WHO guidelines for malaria. Geneva: World Health Organization. [VERIFY: year of edition used]

\bibitem{who2022tb} World Health Organization. WHO consolidated guidelines on tuberculosis, Module 4: treatment. Geneva: World Health Organization. [VERIFY: year of edition used]

\bibitem{ollama} Ollama: run large language models locally. Available from: \url{https://ollama.com}

\bibitem{repo} Oladimeji Gabriel A, Olawuyi D, Ajayi T, Aderemi T. Hausa health drift: benchmark, data and analysis code. Available from: \url{https://github.com/anthoniooladimeji11-coder/hausa-health-drift}

\end{thebibliography}

\end{document}